\documentclass{article}

\usepackage{microtype}
\usepackage{graphicx}
\usepackage{subfigure}
\usepackage{booktabs} 

\usepackage{hyperref}

\usepackage{amsmath} 
\usepackage{amssymb}
\usepackage[T1]{fontenc}

\renewcommand{\cal}{\mathcal} 
\newcommand{\e}{{\varepsilon}}

\newcommand{\E}{\mathbb{E}}

\newcommand{\deq}{\mathrel{\mathop:}=} 

\newcommand{\bfx}{\mathbf{x}}
\newcommand{\bfy}{\mathbf{y}}
\newcommand{\bff}{\mathbf{f}}

\DeclareMathOperator{\Cat}{Cat}
\DeclareMathOperator{\softmax}{softmax}

\newcommand{\eq}[1]{\begin{equation}#1\end{equation}}

\newcommand{\al}[1]{\begin{align}#1\end{align}}

\newcommand{\pa}[1]{\left({#1}\right)}

\newcommand{\qa}[1]{\left[{#1}\right]}

\usepackage[accepted]{icml2020}

\icmltitlerunning{Cold Posteriors and Aleatoric Uncertainty}

\begin{document}

\twocolumn[
\icmltitle{Cold Posteriors and Aleatoric Uncertainty}



\icmlsetsymbol{equal}{*}

\begin{icmlauthorlist}
\icmlauthor{Ben Adlam}{g,ai}
\icmlauthor{Jasper Snoek}{g}
\icmlauthor{Samuel L. Smith}{dm}
\end{icmlauthorlist}

\icmlaffiliation{g}{Google Brain}
\icmlaffiliation{dm}{DeepMind}
\icmlaffiliation{ai}{Work done as a member of the Google AI Residency program (https://g.co/airesidency)}

\icmlcorrespondingauthor{Ben Adlam}{adlam@google.com}

\icmlkeywords{Machine Learning, ICML}

\vskip 0.3in
]



\printAffiliationsAndNotice{}  

\begin{abstract}
Recent work has observed that one can outperform exact inference in Bayesian neural networks by tuning the ``temperature'' of the posterior on a validation set (the ``cold posterior'' effect). To help interpret this phenomenon, we argue that commonly used priors in Bayesian neural networks can significantly overestimate the aleatoric uncertainty in the labels on many classification datasets. This problem is particularly pronounced in academic benchmarks like MNIST or CIFAR, for which the quality of the labels is high. For the special case of Gaussian process regression, any positive temperature corresponds to a valid posterior under a modified prior, and tuning this temperature is directly analogous to empirical Bayes. On classification tasks, there is no direct equivalence between modifying the prior and tuning the temperature, however reducing the temperature can lead to models which better reflect our belief that one gains little information by relabeling existing examples in the training set. Therefore although cold posteriors do not always correspond to an exact inference procedure, we believe they may often better reflect our true prior beliefs.
\end{abstract}

\section{Introduction}
\label{sec_intro}

Bayesians distinguish primarily between two sources of uncertainty, epistemic and aleatoric \citep{der2009aleatory, kendall2017uncertainties}. Epistemic uncertainty is the uncertainty that arises because we have limited training data, or because our prior knowledge regarding the data-generation process is weak. In essence, it arises because many model functions are consistent with both the data we have observed and our prior knowledge. Given more prior knowledge or more training data, the epistemic uncertainty can be reduced, which improves the accuracy of our predictions.

\begin{figure}
    \centering
    \includegraphics[width=\linewidth]{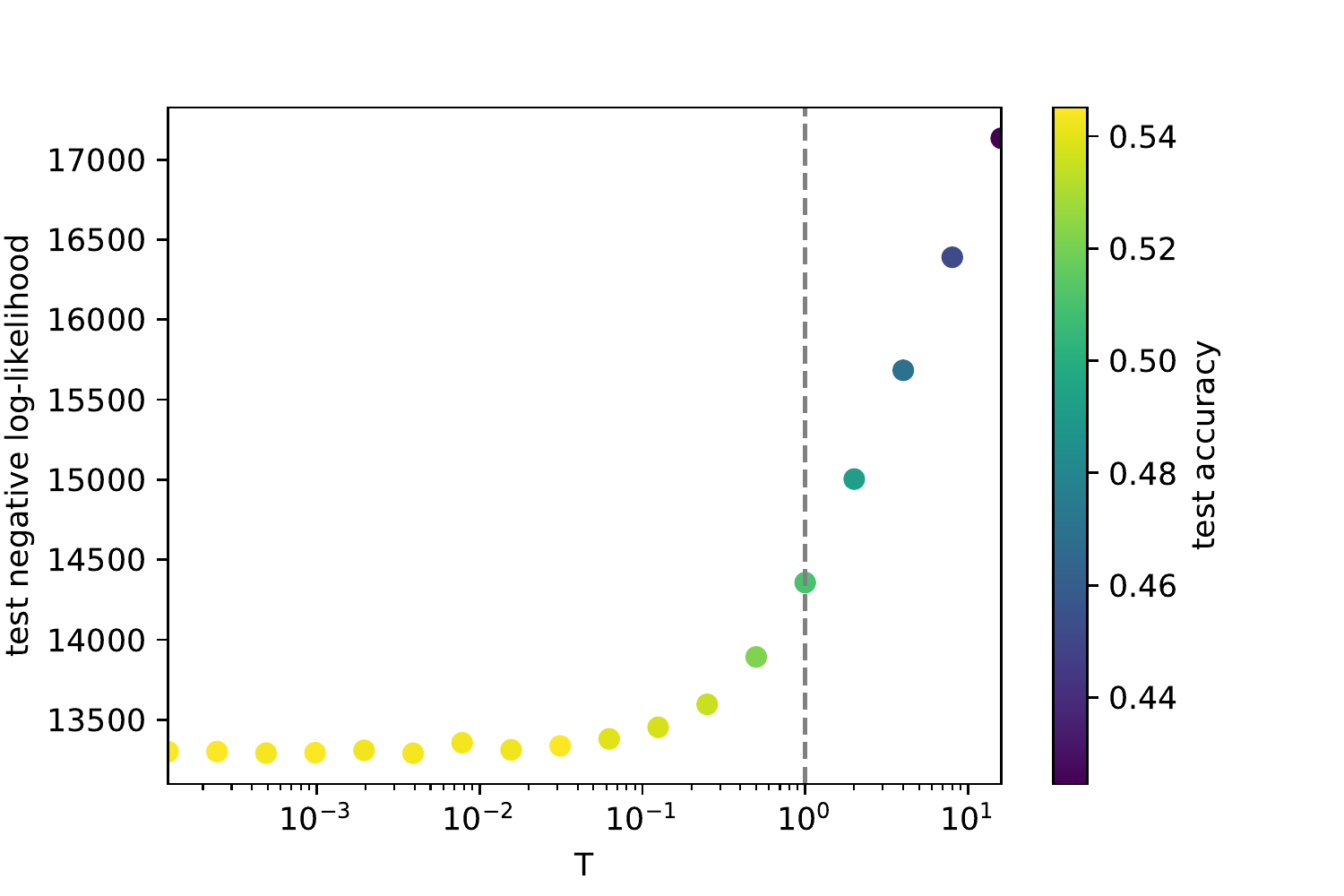}
    \vskip -2mm
    \caption{The cold posterior effect arises in classification problems for Gaussian processes (GPs). We use the model \eqref{eq_gpc} for different temperatures $T$, training on 45K points from CIFAR10. We calculate the log-likelihood and top-1 accuracy on a test set of 10K points. As previously observed for BNNs, we find that temperatures below 1 improve both log-likelihood and accuracy.}
    \label{fig_tempered_gpc}
\end{figure}

Meanwhile, aleatoric uncertainty is the uncertainty that arises from the inherent stochasticity within the data-generation process. Even if we repeat exactly the same experiment under identical conditions, there is a chance that we may obtain different results. For instance, in the context of image classification, if we show exactly the same image to two human labellers (selected randomly from some population), there is some probability that they may assign different labels. No matter how much data we collect, the aleatoric uncertainty will always be present.

When we specify a prior over functions, this prior should capture both the epistemic and the aleatoric uncertainty. If the epistemic uncertainty is high, then the prior will have support across a wide range of different candidate functions. In the presence of aleatoric uncertainty, these functions are stochastic processes, which assign non-zero probability to multiple labels for any given input. However if the aleatoric uncertainty in the data-generation process is low, then the candidate functions under the prior should assign high probabilities (close to 1) to a single label for any given input.  

\begin{figure*}[!ht]
    \centering
    \subfigure[]{\includegraphics[width=0.4\linewidth]{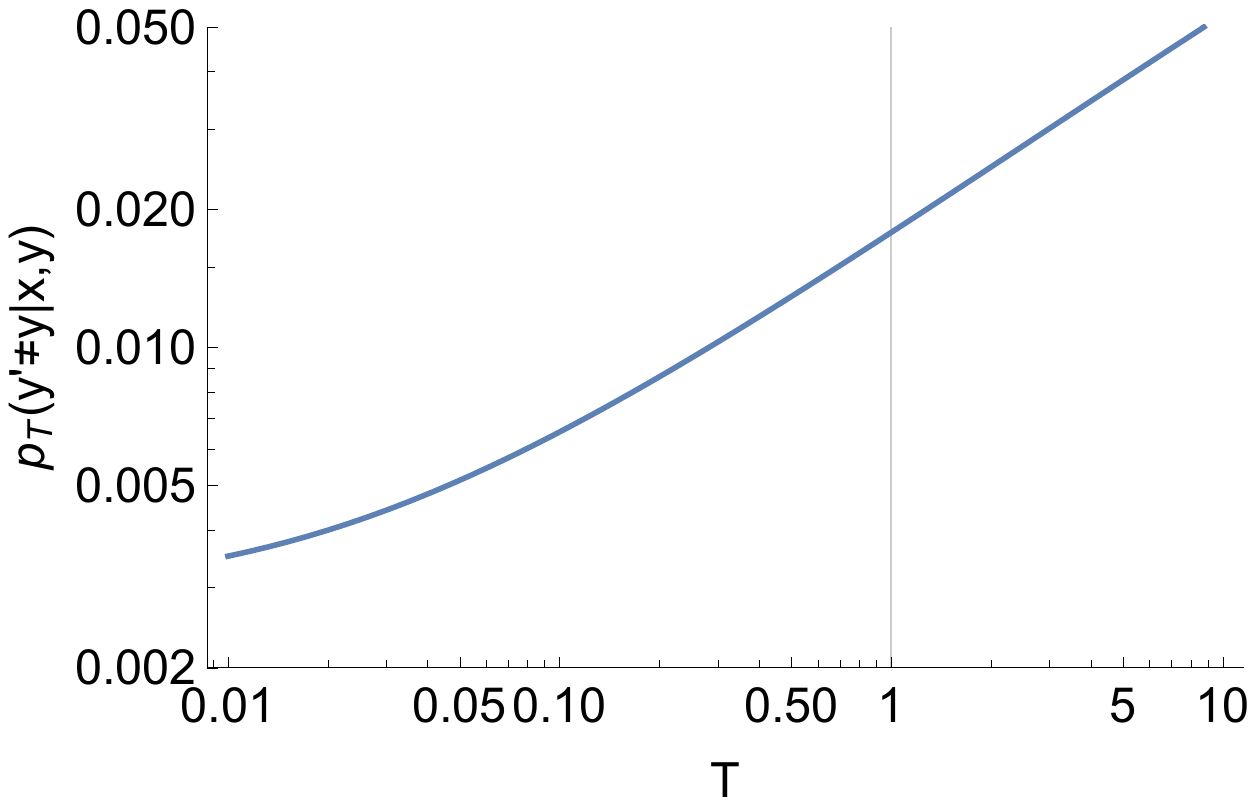}}
    \subfigure[]{\includegraphics[width=0.5\linewidth]{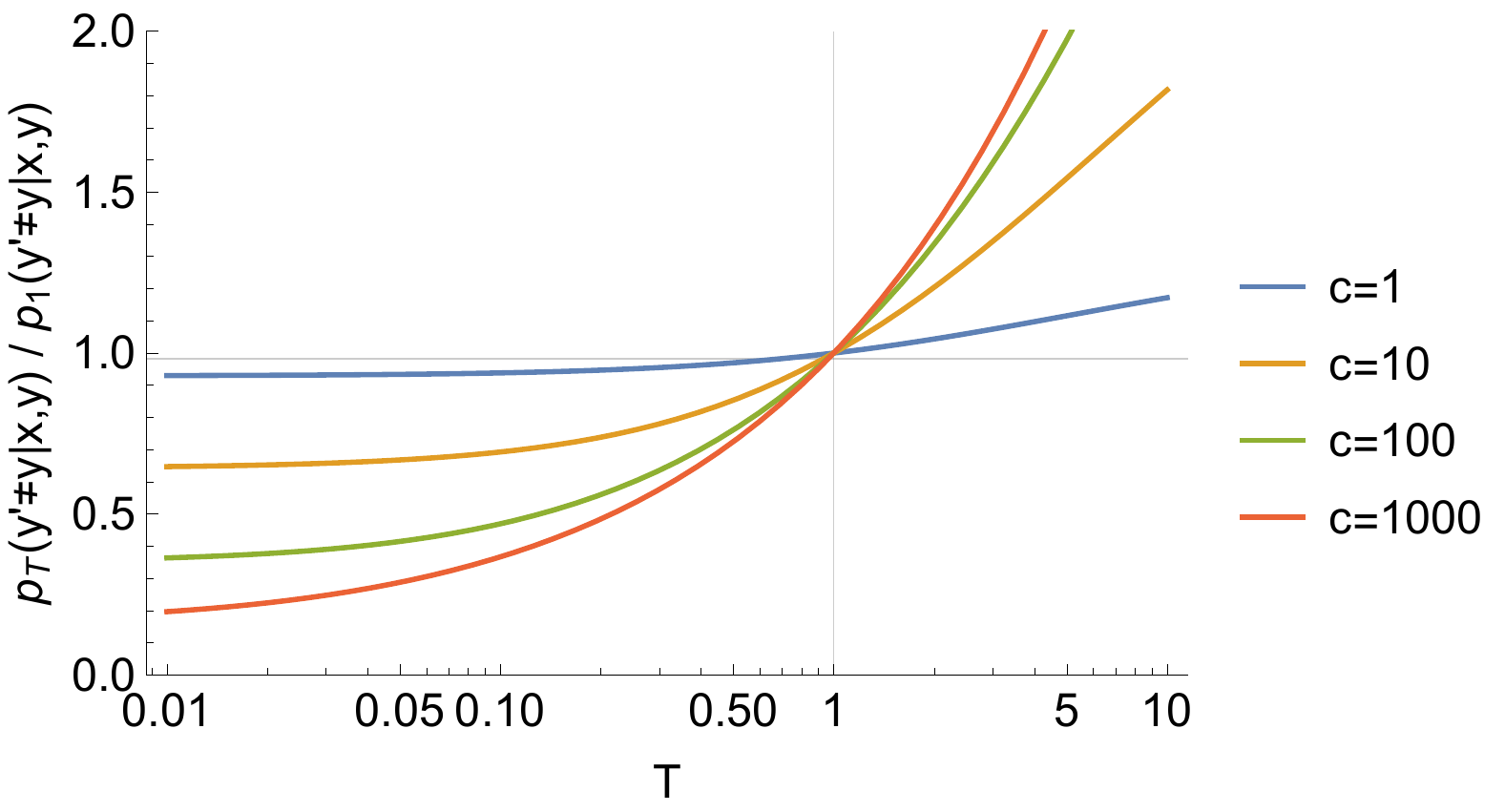}}
    \vskip -2mm
    \caption{Tempering the posterior of GP classification can significantly reduce the aleatoric uncertainty implied by the prior. We consider binary classification under the model (\ref{eq_gpc}), which we introduce in Sec.~\ref{sec:two}. For simplicity, we directly impose a prior on the latent space, $f(x)\sim\mathcal{N}({\bf 0}, cI)$ where $I$ denotes the identity matrix (this is equivalent to studying a GP prior on inputs). (a) When the latent space scale $c=1000$, at temperature $T=1$ the prior assigns a $2\%$ probability that two labels $y$ and $y'$ of a randomly selected input $x$ will differ, while this probability falls to $\sim 0.4\%$ at $T=0.01$. (b) Consider the ratio of the probabilities that two labels of the same input differ for the tempered posterior divided by the Bayesian posterior. We plot this for a range of latent space scales $c$ (increasing $c$ corresponds to reducing the $L_2$ coefficient in BNNs). For any given scale $c$, the aleatoric uncertainty falls as the temperature decreases, before reaching an asymptote at $T=0$. As the scale $c$ increases, the value of the asymptote approaches 0.}
    \label{fig_tempered_aleatoric }
\end{figure*}

Intuitively, we would expect the aleatoric uncertainty in a dataset to be high if either the inputs are compatible with multiple labels, or if the labels provided are unreliable. We note that on some popular academic datasets like MNIST or CIFAR10, neither of these possibilities arise \citep{lecun1998gradient, krizhevsky2009learning}. When these datasets are collected, an effort is usually made to filter out inconclusive images (such as heavily occluded objects), and the quality of the labels are often carefully verified. If I told you that a MNIST digit was a ``7'' but did not show you the image, you would be very confident that the digit was indeed a seven. This contrasts strongly with many large commercial datasets, where the labels are often inferred from user behaviour.

Bayesian neural networks (BNNs) are a natural way to deal with the epistemic uncertainty in over-parameterized neural networks (NNs), as they formally specify a prior over the parameters \citep{mackay1995probable}. However researchers do not always verify that the priors used in BNNs are compatible with our prior beliefs regarding the data-generation process we wish to model. Instead, these priors are often chosen to be consistent with the initialization schemes used when training vanilla NNs. These initialization schemes have been selected over time, largely by trial and error, to facililate efficient optimization \citep{sutskever2013importance, hanin2018start}. Recently, \citet{wenzel2020good} provided empirical evidence to suggest that one can often achieve lower test log-likelihoods and higher test accuracies in BNNs by tempering the posterior to be ``colder'' than the Bayesian posterior. If we let $p(\theta|X,\bff)$ be the Bayesian posterior over parameters, which can be viewed as a Gibbs measure w.r.t. to an energy function $U(\theta)$, \emph{i.e.} $p(\theta|X,\bff) \propto \exp(-U(\theta))$. Then tempering corresponds to performing inference by averaging over parameters drawn from $p(\theta|X,\bff) \propto \exp(-U(\theta)/T)$ for a temperature $T$ not equal to 1. We verify this phenomenon for GP classification on CIFAR10 in Fig.~\ref{fig_tempered_gpc}.


The Bayesian posterior must be optimal if the model is well-specified. The observation of a cold posterior effect therefore indicates either that we are failing to perform accurate inference or that the prior imposed by existing BNNs is poorly chosen. In this paper, we demonstrate that the prior over functions associated with naively chosen parameter priors can have surprisingly high aleatoric uncertainty on classification tasks. Interestingly, tempering the posterior usually reduces this aleatoric uncertainty. For clarity, we consider BNNs in the infinite-width limit, where the prior over function space is described by a GP with a particular kernel, which we refer to as the Neural Network Gaussian Process (NNGP) \cite{neal1996priors, lee2017deep, matthews2018gaussian}. To build our intuition, we also discuss GP regression, for which it is possible to perform exact inference at any temperature. In this simple case, any tempered posterior corresponds to valid Bayesian inference under a modified prior, and tuning the temperature on the validation set is therefore directly equivalent to Empirical Bayes.

\section{The Cold Posterior Effect can be Observed in Gaussian Process Classification}
\label{sec:two}

\citet{neal1996priors} identified a powerful connection between infinite-width NNs and GPs, showing that the outputs of a random, independently initialized single-hidden-layer NN converges to a GP as the number of neurons in the hidden layer approaches infinity. \citet{lee2017deep} extended this result to deep NNs. Let $z_i^l(x)$ describe the $i$th pre-activation following a linear transformation in the $l$th layer of a NN. At initialization, the parameters of the NN are independent and random, so the central-limit theorem can be used to show that the pre-activations become Gaussian with zero mean and a covariance matrix $K(x, x') =  \E[z^l_i(x)z^l_i(x')]$. We use the Neural Tangents library to compute these covaraince matrices \cite{novak2019neural}.
\begin{figure*}[!ht]
    \centering
    \subfigure[]{\includegraphics[width=0.45\linewidth]{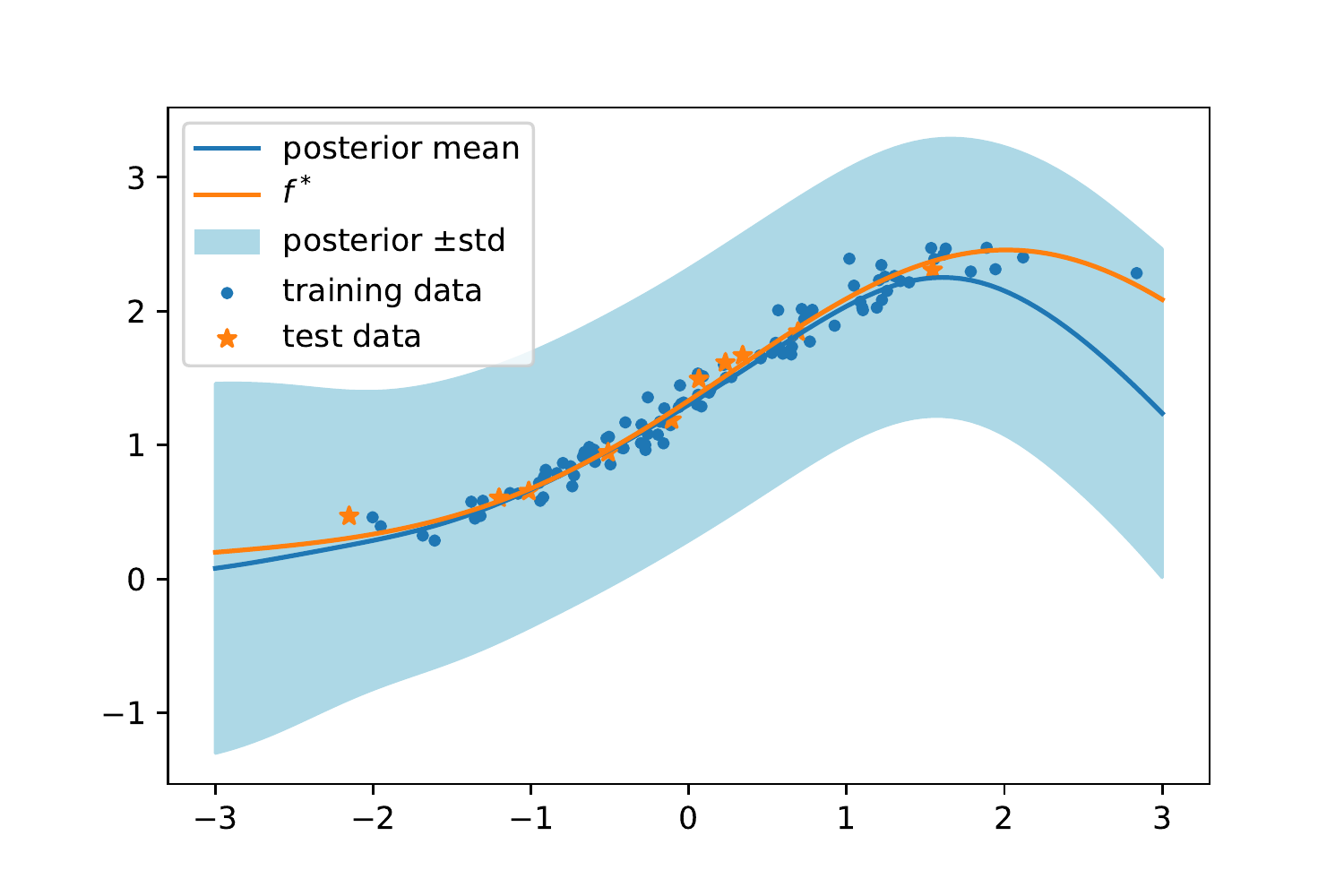} \label{fig:3a}}
    \subfigure[]{\includegraphics[width=0.45\linewidth]{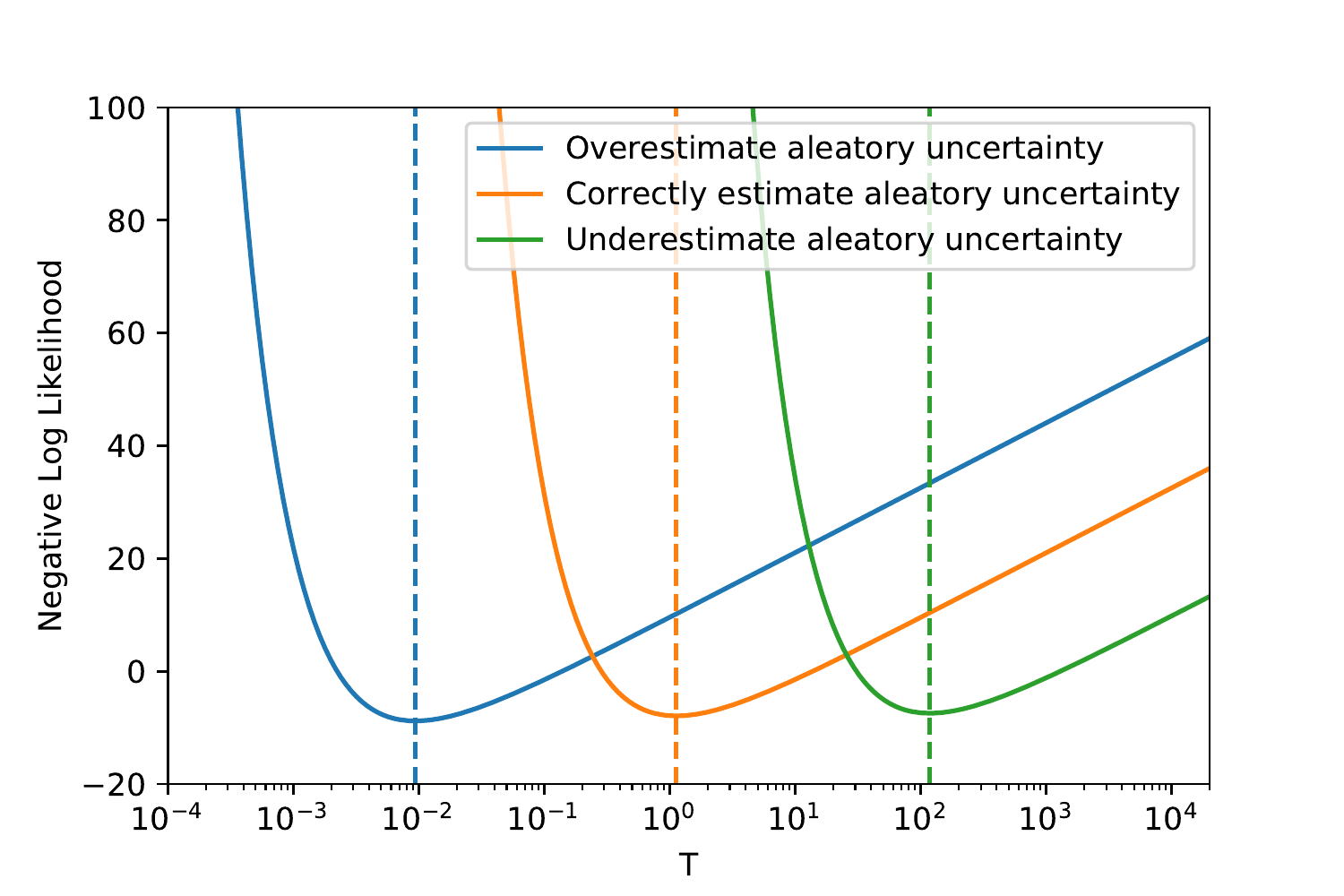} \label{fig:3b}}
    \vskip -2mm
    \caption{We perform GP regression using the same kernel used to generate the data, but for three different estimates of the scale of  aleatoric uncertainty $\sigma_{\e}$. (a) In this figure, we over-estimate the scale of the aleatoric uncertainty. We find that for most test inputs, the mean function is significantly more accurate than the variance associated with the Bayesian posterior would suggest. (b) In this figure, we evaluate the negative log-likelihood at as a function of temperature. If we estimate the aleatoric uncertainty correctly, then the log likelihood is minimized at $T=1$. However if we over-estimate the aleatoric uncertainty, then the log likelihood is minimized for $T < 1$, while if we underestimate the aleatoric uncertainty the log likelihood is minimized for $T > 1$.}
    \label{fig_gpr}
\end{figure*}

We showed in Fig.~\ref{fig_tempered_gpc} that the cold posterior effect can be replicated in classification tasks with the NNGP. Specifically, we consider the model,
\eq{\label{eq_gpc}
    (y|f)\sim \Cat(\softmax(f(\mathbf{x}))),
}
where $f$ is a GP with mean 0 and kernel $K$ defined above.\footnote{We use a ``critically initialized'' NNGP with two hidden layers.} Since the posterior of GP classification is intractable \cite{williams2006gaussian}, we perform inference using elliptical slice sampling \cite{murray2010elliptical} to marginalize over the latent space. Denote the training points as $(\bfx_i, y_i)$ for $i\in\{1,\ldots,n\}$, and collect these points as matrices $X$ and $\bfy$. The tempered posterior for the latent space of a test point $x^*$, denoted $f^*$, is 
\eq{\label{eq_tempering_gpc}
    p_\textsc{t}(f^* | X, \bfy) \propto \E_F\qa{ \prod_i\softmax(f_i)_{y_i}^{1/T} \phi_{\textsc{t}}(f^*\hspace{-.1em}, F) },
}
where $f_i\deq f(\bfx_i)$, $F$ is the matrix containing all $f_i$, and $\phi_{\textsc{t}}(f^*, F)\equiv \phi_{\textsc{t}}(f^*, f_1,\ldots,f_n)$ is the p.d.f. of a multivariate Gaussian
\eq{
     \cal{N}\pa{\mathbf{0}, T\begin{pmatrix} K(\bfx^*,\bfx^*) & K(\bfx^*,X) \\ K(X, \bfx^*) & K(X,X) \end{pmatrix}}.
}

\paragraph{Uncertainty implications of tempering.} In model \eqref{eq_gpc}, tempering has interesting implications for the aleatoric  uncertainty that our model implies. As discussed in the introduction, we can probe this by considering the posterior on the label of a point $x$, which we denote with $y'$, for which we have already observed the label $y$. Specifically, we can ask for the probability that the new label will differ:
\eq{
    p(y'\neq y|x,y) = \E_f\qa{ \sum_{y'\neq y}\hspace{-0.1em}\softmax(f)_{y'} p(f|x,y)},
}
where $p(f|x,y)$ is the posterior for the latent space of $x$. By tempering, as in \eqref{eq_tempering_gpc}, we can understand $p_\textsc{t}(y'\neq y|x,y)$ as a function of $T$ (see Fig.~\ref{fig_tempered_aleatoric }). We find that reducing the temperature $T$ consistently reduces the probability that two labels $y$ and $y'$ drawn at the same point $x$ will differ. 

\textbf{Inference in the absence of aleatoric uncertainty.} Consider the extreme case, in which there is no aleatoric uncertainty. In this case, we require that $p(y'\neq y|x,y) = 0$. To achieve this, all functions in the posterior must interpolate the training data. It is interesting to note that this criterion, which is not satisfied by existing BNNs with standard priors \citep{wenzel2020good}, is often satisfied by the naive MAP ensembling techniques popular in the deep learning community \citep{Lakshminarayanan2017, wilson2020case}, which ensemble over multiple minima identified using different random initializations. Should we be surprised that these simple techniques are often better than existing BNNs at modelling uncertainty on datasets with reliable labels?

\section{Gaussian Process Regression}



Unfortunately the classification case \eqref{eq_gpc} is difficult to study since exact inference is not tractable. To further develop our intuition, we consider the simpler case of GP regression \cite{williams2006gaussian}. We model,
\eq{\label{eq_gpr}
    y\sim f(x) ,
}
where $f$ is again a GP. The posterior on a test point $x^*$ is $ y^*|(x^*, X,\bfy) \sim N(\mu, \sigma^2)$ for mean and variance,
\al{\label{eq_gpr_post}
   \mu &= K(x^*,X)K(X,X)^{-1}\bfy \quad\text{and}\\
   \sigma^2 &= K(x^*,x^*) - K(x^*,X) K(X,X)^{-1}K(X,x^*)\nonumber.
}
In this simple case, tempering the posterior to temperature $T$ also has a closed-form solution, resulting in a predictive function $y|(x^*,X,\bfy) \sim N(\mu_\textsc{t}, \sigma_\textsc{t}^2)$ with the same mean $\mu_\textsc{t} = \mu$ but with variance $\sigma_\textsc{t}^2 = T\sigma^2$. This follows directly from the p.d.f. of a Gaussian and the definition of tempering in Sec.~\ref{sec_intro}. We observe that the following are equivalent:
\begin{itemize}
    \item The Bayesian posterior for the kernel $T K(\cdot,\cdot)$.\footnote{This follows from the cancellation of both $T$ factors in the mean in Eq.~\eqref{eq_gpr_post}, but the variance retaining a factor of $T$.}
    \item The $T$-tempered posterior for the kernel $K(\cdot,\cdot)$.
\end{itemize}
We can therefore interpret tuning the temperature as being directly analogous to tuning the scale of the kernel $K(\cdot,\cdot)$.

\paragraph{Aleatoric uncertainty for regression.} GP regression under model \eqref{eq_gpr} has no aleatoric uncertainty because the implied correlation between independent labels of a single point $x$ is 1. To introduce aleatoric uncertainty, we model,
\eq{
    (y|f(x)) \sim \mathcal{N}(f(x), \sigma_\e^2),
}
where $\sigma_\e$ specifies the scale of the inherently unpredictable label noise. This modifies the posterior in Eq.~\eqref{eq_gpr_post} to,
\al{\label{eq_gpr_post_eps}
   \mu &= K(x^*,X)\tilde{K}^{-1}\bfy \quad\text{and}\\
   \sigma^2 &= K(x^*,x^*)  - K(x^*,X) \tilde{K}^{-1}K(X,x^*) +\sigma_\e^2\nonumber,
}
where $\tilde{K}\deq K(X,X)+\sigma_\e^2I$. As observed above, the tempered posterior has mean $\mu_{\textsc{t}} = \mu$ and variance $\sigma_{\textsc{t}}^2 = T\sigma^2$. Tempering is therefore equivalent to simultaneously rescaling the kernel $K(\cdot,\cdot) \rightarrow TK(\cdot, \cdot)$ and the aleatoric uncertainty $\sigma_{\e}^2 \rightarrow T\sigma_{\e}^2$. We conclude that performing inference at a temperature $T < 1$ both modifies the prior and also reduces the scale of the aleatoric uncertainty $\sigma_{\e}^2$. We therefore expect that tempering will outperform exact inference if either the kernel is misspecified, or if the scale of aleatoric uncertainty $\sigma_\e^2$ has been estimated incorrectly.

To illustrate this, we consider a simple, 1D synthetic dataset. Let $K$ be a standard RBF kernel. We generate 100 i.i.d. datapoints $x_i\sim\mathcal{N}(0, 1)$ and label them by sampling a true function $f^*$ from the prior \eqref{eq_gpr} and adding a small amount of label noise, setting $\sigma_\e=0.1$, so that $y_i=f^*(x_i)+\e_i$. In Fig.~\ref{fig:3a}, we perform exact inference at temperature $T=1$, but we mistakenly over-estimate the scale of aleatoric uncertainty, setting $\sigma_{\e} = 1$. Unsurprisingly, we find that most test points are significantly closer to the posterior mean than one would expect given the posterior variance.

In Fig.~\ref{fig:3b}, we evaluate the negative log likelihood as a function of the temperature $T$ for three different scenarios. For the blue curve, we overestimate the aleatoric uncertainty ($\sigma_{\e} = 1$), for the orange curve we correctly estimate the aleatoric uncertainty ($\sigma_{\e} = 0.1$), and for the green curve we underestimate the aleatoric uncertainty ($\sigma_{\e} = 0.01$). In each case, we perform inference with the same kernel $K$ used to generate the data. As expected, if the aleatoric uncertainty is estimated correctly, the negative log likelihood is minimized at $T=1$. If we overestimate the aleatoric uncertainty the negative log likelihood is minimized for $T<1$ and if we underestimate the aleatoric uncertainty the negative log likelihood is minimized for $T>1$.

\section{Discussion}

Many popular benchmark datasets have very little aleatoric uncertainty. On such datasets, Bayesian inference should ensemble over a diverse range of functions, all of which are highly confident in the vicinity of the training data. This requirement is satisfied by ensembles of vanilla NNs trained with small $L_2$ coefficients, however it is not necessarily satisfied by BNNs if the priors are not carefully chosen.

Although BNNs provide a principled approach to modelling uncertainty in deep networks, they are not widely used in practice. Ensembles of deep vanilla NNs are considered state of the art for both classification accuracy and uncertainty estimation on in-distribution, out-of-distribution, and shifted datasets \cite{Lakshminarayanan2017,ovadia-19}. We note that \citet{wilson2020case} has argued that model averaging is the key component of Bayesian inference. In this weak sense, ensemble methods could be considered Bayesian as they average over MAP training of the same architecture for several random initializations. 

More principled Bayesian methods are unlikely to outperform such ensembles without good priors. Indeed, bad priors have as much potential to hurt generalization as to help. We argue that one flaw in current BNN priors is that they often imply very high aleatoric uncertainty, and we suggest that the cold posterior effect might be explained in part as a post-hoc method of reducing this source of uncertainty. More broadly, we believe that there is no reason to expect that initialization schemes which achieve good performance in vanilla NNs will give rise to appropriate priors for BNNs.


\bibliography{ref}
\bibliographystyle{icml2020}

\end{document}